\RequirePackage{tikz}
\documentclass[pdflatex,default,iicol]{sn}
\usepackage{graphicx}
\usepackage{xurl}
\usepackage{hyperref}
\usepackage{textcomp} 
\usepackage{ulem} 
\usepackage{tabularx}
\usepackage[font=small,skip=0pt]{caption}
\usepackage[font=small,skip=0pt]{subcaption}
\usepackage{lscape}
\usepackage{tikz}
\usetikzlibrary{decorations.pathreplacing} 
\usetikzlibrary{shapes.geometric}
\usepackage{dblfloatfix} 
\usepackage{grffile} 
\usepackage{float} 
\usepackage{xspace} 
\usepackage{xcolor}
\usepackage{soul}
\usepackage{amsmath}

\hypersetup{
    colorlinks = true,
    linkcolor=red,
    citecolor=red,
    urlcolor=blue,
    linktocpage 
}
\definecolor{revcolor}{RGB}{124,2,170}

\newcommand{\diff}[1]{\noindent{#1}}
\newcommand{\banosdataset}[0]{Banos \textit{et al}\xspace}
\newcommand{\recofitdataset}[0]{Recofit\xspace}
\newcommand{\covtypedataset}[0]{Covtype\xspace}

\newcommand{\mondrianforest}[0]{Mondrian forest\xspace}
\newcommand{\mondrianforests}[0]{Mondrian forests\xspace}
\newcommand{\mondriantree}[0]{Mondrian tree\xspace}
\newcommand{\mondriantrees}[0]{Mondrian trees\xspace}

\tikzset{
		cross/.pic = {
		\draw[rotate = 45] (-#1,0) -- (#1,0);
		\draw[rotate = 45] (0,-#1) -- (0, #1);
		}
}

\begin{document}

\title[Mondrian F]{Mondrian Forest for Data Stream Classification Under Memory Constraints}

\author[1]{\fnm{Martin} \sur{Khannouz}}\email{martin.khannouz@concordia.ca}

\author[1]{\fnm{Tristan} \sur{Glatard}}\email{tristan.glatard@concordia.ca}

\affil*[1]{\orgdiv{Department of Computer Science
and Software Engineering}, \orgname{Concordia
University}, \orgaddress{
\city{Montreal},
\state{Quebec}, \country{Canada}}}

\abstract{
Supervised learning algorithms generally assume the availability of
enough memory  to store data models during the training and test
phases. However, this assumption is unrealistic when
data comes in the form of infinite data streams, or when learning
algorithms are deployed on devices with reduced
amounts of memory. In this
paper, we adapt the online \mondrianforest classification algorithm
to work with memory constraints on data streams. In particular, we design five
out-of-memory strategies to update \mondriantrees with new data points when
the memory limit is reached. Moreover, we design node trimming mechanisms to
make \mondriantrees more robust to concept drifts under memory constraints.
We evaluate our algorithms on a variety of real and simulated datasets, and
we conclude with recommendations on their use in different situations: the
Extend Node strategy appears as the best out-of-memory strategy in all
configurations, whereas different node trimming mechanisms should be adopted
depending on whether a concept drift is expected. All our methods are implemented in the OrpailleCC 
open-source library and are ready to be used on embedded systems and connected objects.
}


\maketitle

\section{Introduction}
\label{sec:introduction}
Supervised classification algorithms mostly assume the availability of
abundant memory to store data and models. This is an issue when processing
data streams --- which are infinite sequences by definition --- or when
using memory-limited devices as is commonly the case in the Internet of
Things. This paper studies how the \mondrianforest, a popular online
classification method, can be adapted to work with data
streams and memory constraints compatible with connected objects.

Although online and data stream classification methods both assume that the dataset is available as a sequence of
elements, data stream methods assume that the dataset is infinite whereas online methods consider that it is large but still 
bounded in size. Consequently, online methods usually store the processed elements for future access whereas data stream 
methods do not.

Under memory constraints, a data stream classification model that
has reached its memory limit faces two issues: (1) how to update the model when
new data points \diff{arrive}, which we denote as the \emph{out-of-memory}
strategy, and (2) how to adapt the model to concept drifts, i.e., changes in the learned
concepts. The mechanisms described in
this paper address these two issues in the \mondrianforest.

The \mondrianforest is a tree-based, ensemble, online learning method with
comparable performance to offline Random Forest~\cite{mondrian2014}. Previous
experiments highlighted \diff{that \mondrianforest are sensitive} to the amount of
available memory~\cite{khannouz2020}, which motivates their extension to
memory-constrained environments. In practice, existing
implementations~\cite{mondrian_implementation_1, mondrian_implementation_2} of
the \mondrianforest all assume enough memory and crash when memory is not
available.

The concept drift is a common problem in data streams that occurs when the
distribution of features changes throughout the stream. By design, the
\mondrianforest is not equipped to adapt to concept drifts as its trees cannot
be pruned, trimmed, or modified. When memory is saturated, this lack of adaptability to concept drift worsens 
as trees cannot even grow new branches to accommodate changes in feature distributions.
Therefore, the \mondrianforest needs a
mechanism to free memory such that new tree nodes can grow. Such a
mechanism might also be useful for stable data streams as it would replace less
accurate nodes with better performing ones.

In summary, this paper makes the following contributions:
\begin{enumerate}
\item We adapt the \mondrianforest for data streams;
\item We propose five new out-of-memory strategies for the \mondrianforest under memory constraints;
\item We propose three new node trimming mechanisms to make \mondrianforest adaptive to concept drifts;
\item We evaluate our strategies on six simulated and real datasets.
\end{enumerate}

\section{Materials and Methods}
All the methods presented in this section are implemented in the OrpailleCC
framework~\cite{OrpailleCC}. The scripts to reproduce our experiments are
available on GitHub at
\url{https://github.com/big-data-lab-team/benchmark-har-data-stream}.

\subsection{Mondrian Forest}
\begin{figure*}
		\begin{centering}
				\begin{tikzpicture}[
								roundnode/.style={circle, draw=green!60, fill=green!5, thick, minimum size=2mm},
						]
						\draw (0,0) -- (5.5, 0);
						\draw (0,0) -- (0, 5);
						\node at (-0.5, 3) {F2};
						\node at (3, -0.5) {F1};
						\fill[black!30!green] (1, 2) circle (1mm);
						\fill[black!30!green] (1.2, 1.6) circle (1mm);
						\fill[black!30!green] (1.3, 1.3) circle (1mm);
						\fill[black!40!blue] (2.0, 0.6) circle (1mm);
						\fill[black!40!blue] (2.1, 1.7) circle (1mm);
						\fill[black!40!blue] (2.4, 1.35) circle (1mm);

						\fill[black!30!green] (2.05, 2.6) circle (1mm);
						\fill[black!30!green] (2.08, 3.6) circle (1mm);
						\fill[black!30!green] (3.2, 4.0) circle (1mm);
						\fill[black!30!green] (3.5, 3.0) circle (1mm);

						\fill[black!40!blue] (3.8, 1.95) circle (1mm);
						\fill[black!30!green] (4.2, 1.1) circle (1mm);

						\fill[black!40!blue] (4.3, 4.1) circle (1mm);
						\fill[black!40!blue] (4.9, 3.0) circle (1mm);
						\fill[black!40!blue] (5.3, 1.95) circle (1mm);
						\fill[black!40!blue] (5.0, 0.7) circle (1mm);
						\fill[black!30!green] (4, 3.5) circle (1mm);

						\draw (3.1,2.9) -- (3.1, 4.2);
						\draw (5.0,2.9) -- (5.0, 4.2);
						\draw (3.1,4.2) -- (5.0, 4.2);
						\draw (3.1,2.9) -- (5.0, 2.9);
            \node at (3.42, 4.05) {D};

						\draw (3.7,0.6) -- (3.7, 2.05);
						\draw (5.4,0.6) -- (5.4, 2.05);
						\draw (3.7,2.05) -- (5.4, 2.05);
						\draw (3.7,0.6) -- (5.4, 0.6);
            \node at (3.85, 1.7) {C};

						\draw (3.08,0.58) -- (3.08, 4.22); 
						\draw (5.42,0.58) -- (5.42, 4.22);
						\draw (3.08,4.22) -- (5.42, 4.22);
						\draw (3.08,0.58) -- (5.42, 0.58);
						\draw [dashed] (3.08,2.6) -- (5.42, 2.6);
            \node at (3.28, 2.45) {B};

						\draw (0.9,0.5) -- (0.9, 3.7);
						\draw (2.5,0.5) -- (2.5, 3.7);
						\draw (0.9,3.7) -- (2.5, 3.7);
						\draw (0.9,0.5) -- (2.5, 0.5);
            \node at (1.2, 3.5) {A};

						\draw (0.88,0.48) -- (0.88, 4.24);
						\draw (5.44,0.48) -- (5.44, 4.24);
						\draw (0.88,4.24) -- (5.44, 4.24);
						\draw (0.88,0.48) -- (5.44, 0.48);
						\draw [dashed] (2.6,0.48) -- (2.6, 4.24);
            \node at (1, 4.4) {R};

						\node[roundnode] at (8.5, 4) (root) {R};
						\node[roundnode] at (7.5, 2.5) (a) {A};
						\draw[->] (root.south west) -- (a.north);
						\node[roundnode] at (9.5, 2.5) (b) {B};
						\draw[->] (root.south east) -- (b.north);

						\node[roundnode] at (8.5, 1.3) (c) {C};
						\draw[->] (b.south west) -- (c.north east);
						\node[roundnode] at (10.5, 1.3) (d) {D};
						\draw[->] (b.south east) -- (d.north west);

						\fill[purple!60] (3.5, 1.8) circle (1mm);

						\draw [dashed] (-1,-1.2) -- (12, -1.2);

						\draw (0,-7) -- (5.5, -7);
						\draw (0,-7) -- (0, -2);
						\node at (-0.5, -4) {F2};
						\node at (3, -7.5) {F1};

						\fill[black!30!green] (1, -5) circle (1mm);
						\fill[black!30!green] (1.2, -5.4) circle (1mm);
						\fill[black!30!green] (1.3, -5.7) circle (1mm);
						\fill[black!40!blue] (2.0, -6.4) circle (1mm);
						\fill[black!40!blue] (2.1, -5.3) circle (1mm);
						\fill[black!40!blue] (2.4, -5.65) circle (1mm);

						\fill[black!30!green] (2.05, -4.4) circle (1mm);
						\fill[black!30!green] (2.08, -3.4) circle (1mm);
						\fill[black!30!green] (3.2, -3) circle (1mm);
						\fill[black!30!green] (3.5, -4) circle (1mm);

						\fill[black!40!blue] (3.8, -5.05) circle (1mm);
						\fill[black!30!green] (4.2, -5.9) circle (1mm);

						\fill[black!40!blue] (4.3, -2.9) circle (1mm);
						\fill[black!40!blue] (4.9, -4) circle (1mm);
						\fill[black!40!blue] (5.3, -5.05) circle (1mm);
						\fill[black!40!blue] (5.0, -6.3) circle (1mm);
						\fill[black!30!green] (4, -3.5) circle (1mm);

						\draw (3.1,-4.1) -- (3.1, -2.8);
						\draw (5.0,-4.1) -- (5.0, -2.8);
						\draw (3.1,-2.8) -- (5.0, -2.8);
						\draw (3.1,-4.1) -- (5.0, -4.1);
            \node at (3.42, -2.95) {D};

						\draw (3.7,-6.4) -- (3.7, -4.95);
						\draw (5.4,-6.4) -- (5.4, -4.95);
						\draw (3.7,-4.95) -- (5.4,-4.95);
						\draw (3.7,-6.4) -- (5.4, -6.4);
            \node at (3.85, -5.3) {C};

						\draw (3.4,-5.3) -- (3.4, -5.1);
						\draw (3.6,-5.3) -- (3.6, -5.1);
						\draw (3.4,-5.1) -- (3.6, -5.1);
						\draw (3.4,-5.3) -- (3.6, -5.3);

						\draw (3.08,-6.42) -- (3.08, -2.78); 
						\draw (5.42,-6.42) -- (5.42, -2.78);
						\draw (3.08,-2.78) -- (5.42, -2.78);
						\draw (3.08,-6.42) -- (5.42, -6.42);
						\draw [dashed] (3.08,-4.4) -- (5.42, -4.4);
            \node at (3.28, -4.55) {B};

						\draw (0.9,-6.5) -- (0.9, -3.3);
						\draw (2.5,-6.5) -- (2.5, -3.3);
						\draw (0.9,-3.3) -- (2.5, -3.3);
						\draw (0.9,-6.5) -- (2.5, -6.5);
            \node at (1.2, -3.5) {A};

						\draw (0.88,-6.52) -- (0.88, -2.76);
						\draw (5.44,-6.52) -- (5.44, -2.76);
						\draw (0.88,-2.76) -- (5.44, -2.76);
						\draw (0.88,-6.52) -- (5.44, -6.52);
						\draw [dashed] (2.6,-6.52) -- (2.6, -2.76);
            \node at (1, -2.59) {R};

						\node[roundnode] at (8.5, -3) (root1) {R};
						\node[roundnode] at (7.5, -4.5) (a1) {A};
						\draw[->] (root1.south west) -- (a1.north);
						\node[roundnode] at (9.5, -4.5) (b1) {B};
						\draw[->] (root1.south east) -- (b1.north);

						\node[roundnode] at (10.5, -5.7) (d1) {D};
						\draw[->] (b1.south east) -- (d1.north west);

						\node[roundnode, draw=red!60, fill=red!5] at (8.5, -5.7) (e1) {E};
						\draw[->] (b1.south west) -- (e1.north east);
						\node[roundnode] at (9.5, -6.9) (c1) {C};
						\draw[->] (e1.south east) -- (c1.north west);
						\node[roundnode, draw=purple!60, fill=purple!5] at (7.5, -6.9) (f1) {F};
						\draw[->] (e1.south west) -- (f1.north east);

						\draw [red] (3.38,-6.42) -- (3.38, -4.93);
						\draw [red] (5.42,-6.42) -- (5.42, -4.93);
						\draw [red] (3.38,-4.93) -- (5.42, -4.93);
						\draw [red] (3.38,-6.42) -- (5.42, -6.42);
						\draw [red, dashed] (3.65,-6.42) -- (3.65, -4.93);
            \node [purple] at (3.53, -5.5) {F};
            \node [red] at (4.5, -4.75) {E};

						\fill[purple!60] (3.5, -5.2) circle (1mm);
				\end{tikzpicture}
    \caption{\diff{Illustration of the \mondriantree before (top) and after (bottom)
		of a branch out. The left side shows the data points in a two-feature space with
		the node's boxes and the splits. The right side shows the tree structure. The
		branch out is triggered by the purple data
    point.}}
		\label{fig:branch-out}
		\vspace{-0.3cm}
		\end{centering}
\end{figure*}
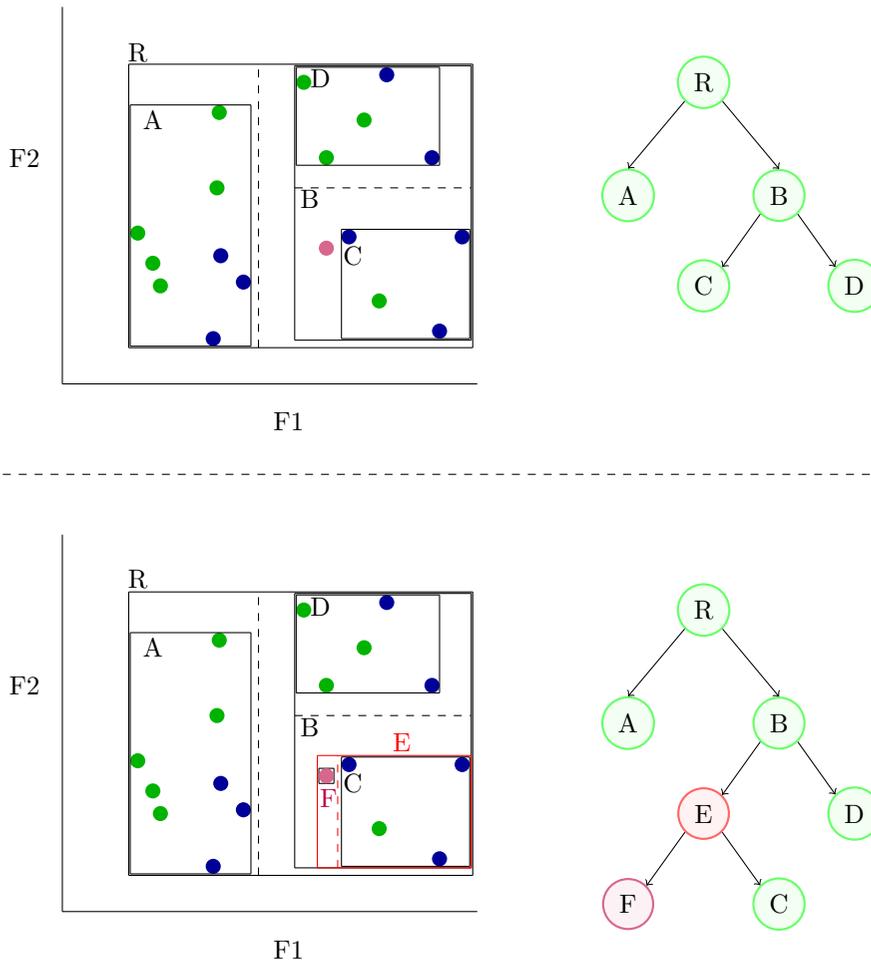

\diff{The \mondrianforest~\cite{mondrian2014} is an ensemble method that
aggregates Mondrian trees. Each tree in the forest
recursively divides the feature space, similar to
a traditional decision tree. However, the feature
selected for the split and its corresponding value
are chosen randomly rather than through an information gain metric. The probability to select
a feature is based on the feature range, and the split
value is uniformly chosen within the feature
range. Unlike other decision trees, the Mondrian
tree does not split leaves to introduce new nodes.
Instead, it adds a new parent and sibling to the
node where the split occurs, using the branch-out
mechanism shown in Figure~\ref{fig:branch-out}. The original node and
its descendants remain unchanged, and only the
data point that initialized the split is moved to
the new sibling. The new parent contains a split
that separates the original node and the new
sibling. This approach enables the Mondrian tree
to introduce new branches to internal nodes. The
branch-out mechanism is randomly triggered when a data point
falls outside of the node box, with a probability related to
 the distance between the new point and the node box.
 The training algorithm does not
require labels to build the tree; however, each
node maintains counters for each observed label.
Therefore, labels can be delayed, but they are
necessary for prediction. Furthermore, each node
tracks the range of its feature, representing a
box that contains all data points. A data point
can create a new branch only if it belongs to a
node and falls outside its box.}

\subsection{Mondrian Forest for Data Stream Classification}
The implementation of \mondrianforest
presented in~\cite{mondrian2014, mondrian_implementation_1} is online because
trees rely on potentially all the previously seen data points to grow new branches.
To support data streams, the \mondrianforest has to access data points only once as the dataset is assumed to be infinite in size.
Algorithm~\ref{alg:training-node}, adapted from reference~\cite{mondrian2014},
describes our data-stream implementation. Function
\texttt{update\_box} (line 11) updates the node's box using the data point's features.
Function \texttt{update\_counters} (line 12) updates the label counters assigned to the
node. Function \texttt{is\_enough\_memory} (line 2) returns true if there is enough
memory to run \texttt{extend\_tree} (line 7). Function
\texttt{distance\_to\_box} (line 3) compute
the distance between a data point and the node's box and it returns zero if the
data point fall inside the box. Function \texttt{random\_bool} (line 5) randomly select a
boolean that is more likely to be true when the data point is far from the
node's box. This function always return false if the distance is zero. Finally,
function \texttt{extend\_tree} extends the tree by introducing a new parent and
a new sibling to the current node.

To make a prediction, a node assigns a score to each label. This score uses
the node's counters, the parent's scores, and the score generated by an
hypothetical split. The successive scores
of the nodes encountered by the test data point are combined together to create the tree
score. Finally, the scores of the trees are averaged to get the forest's score.
The prediction is the label with the highest score.

Mondrian trees can be tuned using three parameters: the base count, the discount
factor, and the budget. The base count is a default score given to the root's
parent. The discount factor controls the contribution of a node to the score of its children.
A discount factor closer to one
makes the prediction of a node closer to the prediction of its parent. Finally,
the budget controls the tree~depth. A small budget makes nodes virtually closer
to each other, and thus, less likely to introduce new splits.


\begin{algorithm}
\begin{algorithmic}[1]

\Require x = a data point
\Require l = label of the data point
\Require node = current node containing attributes in Table~\ref{tab:node-attributes}

\State Function train\_node(x, l, node):
\State m = is\_enough\_memory()\;
\State dist = node.distance\_to\_box(x)\;
\If{dist is positive}
		\State r = random\_bool(node, node.parent, dist)\;
		\If{r and m}
				\State extend\_tree(node, node.parent, x, l)\;
				\State return\;
		\EndIf
\EndIf
\State update\_box(node, x, r, m)\;
\State update\_counters(node, l, r, m)\;
\If{!node.is\_leaf()}
		\State sv = node.split\_value\;
		\State sf = x[node.split\_feature]\;
		\If{sf $<$ sv}
				\State train\_node(x, l, node.right)\;
		\Else
				\State train\_node(x, l, node.left)\;
		\EndIf
\EndIf
\end{algorithmic}
\caption{Recursive function to train a node in the
		data stream \mondrianforest. This function is
		called on each root node of the forest when a
		new labeled data point arrives. The
		implementation of functions
		\texttt{update\_box} and
		\texttt{update\_counters} called at lines 11
		and 12 vary depending on the out-of-memory
		strategy adopted (see
		Section~\ref{sec:method-xp1}).
		Table~\ref{tab:out-of-memory-strategies}
		summarizes these strategies.
		Table~\ref{tab:node-attributes} describes the
		attributes of the node data structure.}
\label{alg:training-node}
\end{algorithm}

\begin{table}
\begin{center}
\resizebox{\linewidth}{!}{
\begin{tabular}{ r c }
\hline
	Attributes & Description \\ 
\hline
parent         & Node's parent or empty for the root. \\
split\_feature & Feature used for the split.\\
split\_value   & Value used for the split.\\
right          & Right child of the node.\\
left           & Left child of the node.\\
counters       & An array counting labels\\
lower\_bound   & An array saving the minimum value for each feature\\
upper\_bound   & An array saving the maximum value for each feature\\
prev\_lower\_bound & Same as lower\_bound but saving the previous minimum\\
prev\_upper\_bound & Same as upper\_bound but saving the previous maximum\\
\hline
\end{tabular}                                                          
}
\caption{Summary of the node structure used in Algorithm~\ref{alg:training-node}.}
\label{tab:node-attributes}
\end{center}
\end{table}

\diff{\mondriantrees are independent from each
other, which allows the forest to be parallelized.
In our experiments, data points are processed sequentially, but in situations where they arrive in
batches, the tree induction process can be
parallelized at the branch level. As a result,
parallelism would increase as the batch
is sorted down the tree. However, this paper
does not focus on parallelizing the
\mondrianforest, because parallelization often increases memory consumption.}

\subsection{Out-of-memory Strategies in the Mondrian Tree}
\label{sec:method-xp1}
A memory-constrained \mondrianforest needs to determine what to do with
data points when the memory limit is reached. We designed five
out-of-memory strategies for this purpose. These strategies specify how the
statistics, namely the counters and the box limits, should be updated when
training nodes. They are implemented in functions \texttt{update\_box} and
\texttt{update\_counters} called at lines 11 and 12 in
Algorithm~\ref{alg:training-node}. Table~\ref{tab:out-of-memory-strategies}
summarizes these strategies.

The \textbf{Stopped} strategy discards any subsequent data point when the memory
limit is reached. It is the most straightforward method as it only uses the
model created so far. In this strategy, functions \texttt{update\_box} and
\texttt{update\_counters} are
no-ops. The Stopped strategy has the advantage of not corrupting the node's box
with outlier data points that would have required a split earlier in the tree.
However, it also drops a lot of data after the model reached the memory limit. 

The \textbf{Extend Node} strategy disables the creation of new nodes when the
memory limit is reached. Each data point is passed down the tree and no splits
are created. However, the counters and the box of each node are updated.
In Algorithm~\ref{alg:training-node}, functions \texttt{update\_box} and
\texttt{update\_counters} work as if there were no split, thus updating 
counters and nodes as if \texttt{r} was always false. 
Compared to the Stopped method, the Extend Node one includes all the data points in the model. However,
since the tree structure does not change, outlier data points may extensively
increase node's boxes and as a result \mondriantrees tend to have large boxes,
which is 1) detrimental to classification performance, and 2) limits further node
creations in the event that more memory becomes available since the distance to the node's box is unlikely to be positive when
computing $m$ at line 4 of Algorithm~\ref{alg:training-node}.

The \textbf{Partial Update} strategy discards the points that would have created
a split if enough memory was available, and updates the model with the other
ones. This strategy discards fewer data points than in the Stopped method while
it is less sensitive to outlier data points than the Extend Node method. In
Algorithm~\ref{alg:training-node}, functions \texttt{update\_box} and
\texttt{update\_counters} update statistics as in the original implementation,
except when a split is triggered ($r=true$). In that case, all modifications done
previously are canceled. Algorithm~\ref{alg:partial-update-counter}
and~\ref{alg:partial-update-box} describe functions
\texttt{update\_counters} and \texttt{update\_box} in more details.

The \textbf{Count Only} strategy never updates node boxes and simply updates the
counters with every new data point. In Algorithm~\ref{alg:training-node}, the
function \texttt{update\_box} is a no-op, while the function
\texttt{update\_counters} works as in the original implementation. This strategy
is less sensitive to outliers than the Extend Node method and it discards less
data points than Partial Update. However, it may create nodes that count data
points outside of their box, thus nodes that do not properly
describe the data distribution.

The \textbf{Ghost} strategy is similar to Partial Update for data points that do not
create a split ($r=false$). In case of a
split ($r=true$), the data point is dropped, however, in contrast with the
Partial Update method, the changes applied between the node parent and the root
are not canceled. This allows internal nodes to keep some information about data
points that would have introduced a split, which preserves some information from
the discarded data points.

\begin{table}
\begin{center}
\resizebox{\linewidth}{!}{
\begin{tabular}{ r c c c c c c c }
\hline
	Method & \texttt{update\_counters} & \texttt{update\_box} \\ 
\hline
Stopped & --- & --- \\
Extend Node  & Always  & Always\\
Partial Update& Only if no split is triggered  & Only if no split is triggered\\
Count Only    & Always & ---\\
Ghost         & Update until a split is triggered & Update until a split is triggered\\
\hline
\end{tabular}                                                          
}

\caption{Summary of the proposed out-of-memory strategies.
\texttt{update\_counters} and \texttt{update\_box} refer to the functions
in Algorithm~\ref{alg:training-node}. ---: no-op.}
\label{tab:out-of-memory-strategies}
\end{center}
\end{table}

\begin{algorithm}
\begin{algorithmic}[1]
\Require node = current node containing attributes in Table~\ref{tab:node-attributes}
\Require l = label of the data point
\Require r = true if a split has been introduced.
\Require m = true if there is enough memory to run extend\_tree.

\State Function update\_counters(node, l, r, m):
\State m = is\_enough\_memory()\;
\State dist = node.distance\_to\_box(x)\;
\State node.counters[l] += 1
\If{m is false and r is true	}
		\While{node.parent is not root}
				\State node.counters[l] -= 1\;
				\State node = node.parent
		\EndWhile
\EndIf
\end{algorithmic}
\caption{Partial Update algorithm for function \texttt{update\_counters}.}
\label{alg:partial-update-counter}
\end{algorithm}

\begin{algorithm}
\begin{algorithmic}[1]
\Require node = current node containing attributes in Table~\ref{tab:node-attributes}
\Require x = the data point
\Require r = true if a split has been introduced.
\Require m = true if there is enough memory to run extend\_tree.

\State Function update\_box(node, x, r, m):
\ForAll{f $\in$ all features}
		\State node.prev\_lower\_bound[f] = node.lower\_bound[f]\;
		\State node.prev\_upper\_bound[f] = node.upper\_bound[f]\;
		\State node.lower\_bound[f] = min(x[f], node.lower\_bound[f])\;
		\State node.upper\_bound[f] = min(x[f], node.upper\_bound[f])\;	
\EndFor
\If{m is false and r is true	}
		\While{node.parent is not root}
				\ForAll{f $\in$ all features}
						\State node.lower\_bound[f] = node.prev\_lower\_bound[f]\;
						\State node.upper\_bound[f] = node.prev\_upper\_bound[f]\;
				\EndFor
				\State node = node.parent
		\EndWhile
\EndIf
\end{algorithmic}
\caption{Partial Update algorithm for function \texttt{update\_box}.}
\label{alg:partial-update-box}
\end{algorithm}

\subsection{Concept Drift Adaptation for Mondrian Forest under Memory Constraint}
\label{sec:method-xp2}
In this section, we propose methods to adapt \mondrianforests to concept
drifts under memory constraints. We design and compare mechanisms to free up some memory by trimming
tree leaves, and resume the growth of the forest after an out-of-memory
strategy was applied. More specifically, we propose three methods to select
a tree leaf for trimming: (1) Random trimming, (2) Data Count, and (3)
Fading Count. In all cases, the trimming mechanism is called periodically
on all trees when the memory limit is reached. 

The \textbf{Random} method, used as baseline, selects a leaf to trim with
uniform probability. The \textbf{Count} method selects the leaf with the
lowest data point counter: it assumes that leaves with few data points are less critical for the classification 
than leaves with many data points. The
exception to this would be leaves that have been recently created. To
address this issue, the \textbf{Fading count} method applies a fading
factor to the leaf counters: when a new data point arrives in a leaf, the leaf
counter $c$ is updated to $c = 1 + c \times f$, where $f$ is the fading
factor. The other leaves get their counter updated to $c = c \times f$. As
a result, leaves that haven't received data points recently are more prone
to be discarded. 

For all methods, the selected leaf is not trimmed if it contains
more data points than a configurable threshold. This threshold prevents the
trimming of an important leaf. The trimming
mechanism is triggered every hundred data points when the memory limit is
reached.

Once leaves have been trimmed, new nodes are available for the forest to
resume its growth. The forest can extend according to the original
algorithm, meaning that only new data points with features outside of the
node boxes can create a split. However, with the Extend Node and Ghost
strategies, this method creates leaves containing mostly outliers since
node boxes have extensively increased when memory was full. This scenario
is problematic because these new leaves might be even less used than the
trimmed ones. 

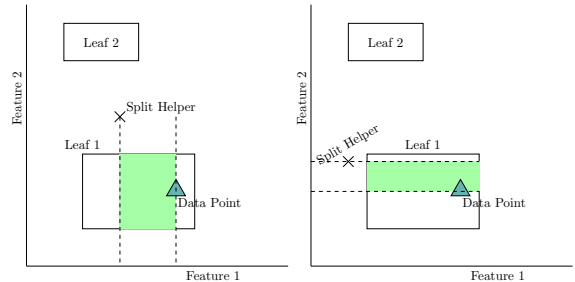
\begin{figure}[!t]
	\centering
	\resizebox{\columnwidth}{!}{
	\begin{tikzpicture}
			\node [draw, line width=0.2mm, minimum width=20mm, minimum height=10mm] at (0, 4) {Leaf 2};
			\node [draw, line width=0.2mm, minimum width=30mm, minimum height=20mm] at (1, 0) {};
			\node [] at (-0.5, 1.25) {Leaf 1};
			\node [minimum width=15mm, minimum height=20mm, fill=green!35] at (1.25, 0) {};
			\draw (0.5,2) pic[rotate=0,black] {cross=2mm};
			\node [] at (1.6, 2.25) {Split Helper};
			\node[isosceles triangle,
			isosceles triangle apex angle=60,
			draw,fill=teal!60,
			minimum size =1pt,
			rotate=90] (T60)at (2,0){};
			\node [] at (2.9, -0.3) {Data Point};

			\draw [dashed] (0.5,2) -- (0.5,-2);
			\draw [dashed] (2,2) -- (2,-2);

			\draw [] (-2,-2) -- (-2,5);
			\draw [] (-2,-2) -- (5,-2);
			\node [rotate=90] at (-2.25, 2.5) {Feature 2};
			\node [] at (3, -2.25) {Feature 1};
	\end{tikzpicture}
	\begin{tikzpicture}
			\node [draw, line width=0.2mm, minimum width=20mm, minimum height=10mm] at (0, 4) {Leaf 2};
			\node [draw, line width=0.2mm, minimum width=30mm, minimum height=20mm] at (1, 0) {};
			\node [] at (1, 1.25) {Leaf 1};
			\node [minimum width=30mm, minimum height=8mm, fill=green!35] at (1, 0.4) {};
			\draw (-1,0.8) pic[rotate=0,black] {cross=2mm};
			\node [rotate=30] at (-1, 1.25) {Split Helper};
			\node[isosceles triangle,
			isosceles triangle apex angle=60,
			draw,fill=teal!60,
			minimum size =1pt,
			rotate=90] (T60)at (2,0){};
			\node [] at (2.9, -0.3) {Data Point};

			\draw [dashed] (-2,0.8) -- (2.5,0.8);
			\draw [dashed] (-2,0) -- (2.5,0);

			\draw [] (-2,-2) -- (-2,5);
			\draw [] (-2,-2) -- (5,-2);
			\node [rotate=90] at (-2.25, 2.5) {Feature 2};
			\node [] at (3, -2.25) {Feature 1};
	\end{tikzpicture}
	}
	\caption{A two-feature example of the split definition. In each scenario, rectangles represent leaf boxes, the triangle
		is the new data point, and the cross is the split helper.
		The green area indicates the range of values where the new split will be created. The
		split helper is arbitrarily placed to illustrate different situations.}
		\label{fig:split-mechanism}
\end{figure}

To address this issue, we propose to split the tree leaves that may have expanded 
as a result of the memory limitation in the Extend Node and Ghost strategies. 
The splitting of tree leaves is defined from two points: a new data
point, and a split helper (Figure~\ref{fig:split-mechanism}). The split is triggered 
in the tree leaf that contains the new data point, along a dimension defined using the split helper. 
We propose two variants of the splitting method corresponding to different definitions of the 
split helpers: in variant \textbf{Split AVG}, the split helper is the fading average data point, whereas 
in \textbf{Split Barycenter}, it is the weighted average of the leaves.
The split dimension is randomly picked amongst the dimensions along which the
split helper features are included within the leaf box. The split value
along this dimension (green region in Figure~\ref{fig:split-mechanism}) is
randomly picked between the data point feature value and the split helper
feature value along this dimension. The counters in the original leaf are proportionally
split between the new leaves.
The idea behind the split helper is to create new leaves in the parts of the tree that already contain
 a lot of data points.

\subsection{Time Complexity}

In this section, we discuss the time complexity of the training and testing
processes. Equations~\ref{eqn:train} and~\ref{eqn:predict} respectively describe
the time complexity for the $training$ and $testing$ processes, where:
\begin{itemize}
		\item $m$ is the memory size in number of nodes.
		\item $t$ is the tree count.
		\item $d$ is the depth of the tree $\simeq log(\frac{m}{t})$ in a case of a
				balanced tree.
		\item $l$ is the label count.
		\item $f$ is the feature count.
\end{itemize}

\begin{equation}
		\label{eqn:train}
		training = \mathcal{O}(t d (f + l))\\
\end{equation}
\begin{equation}
		\label{eqn:predict}
		testing = \mathcal{O}(m l + t l + t d ( f + l) + l) \\
\end{equation}

Indeed, in the training process, the $t$ term is related to training each tree of the
ensemble. The data point is sorted into a leaf, which gives the $d$ term. Finally,
at each depth level, the process computes distances, and updates the node's bounds and
the node counters, which gives us the $(f+l)$ terms.

For the testing process, the model starts by updating statistics in each node
for the labels, which gives $ml$ in Equation~\ref{eqn:predict}. Then it processes the score of
each tree, which gives the $t l$ term. Similarly to the training process, the
predict process will sort the data point into a leaf while computing distances
and nodes score, which adds the term $t d ( f + l )$. Finally, the tree scores
are aggregated and it adds $l$.

We note that, despite being constant in regards to the stream size, the time
complexity is impacted by dataset characteristics (label count and feature
count) as well as user-defined parameters (memory size and tree count). We also
note that for all variables included in Equation~\ref{eqn:train}
and~\ref{eqn:predict}, none of them expanded into quadratic terms.

The out-of-memory strategies do not influence these equations, however, the
trimming and split methods add a few terms to the training process. The
Trim Fading method needs to fade the count of all leaves, which adds $ml$ as
shown in Equation~\ref{eqn:train-trim-fading}.

\begin{equation}
		\label{eqn:train-trim-fading}
		training\_trim\_fading = \mathcal{O}( m l + t d (f + l))\\
\end{equation}

Besides, the Split AVG has to maintain an average data point, which introduces
the term $f$ and gives Equation~\ref{eqn:train-split-avg}.

\begin{equation}
		\label{eqn:train-split-avg}
		training\_split\_AVG = \mathcal{O}( f + t d (f + l))\\
\end{equation}

\subsection{Node Boxes Analysis}
In this section, we analyze the impact of node boxe sizes on the quality of tree
predictions based on the \mondrianforest algorithm provided
in~\cite{mondrian2014}.

In the classification step, the \mondriantree passes the data point $x$ through
the tree, and computes a score $S_k$ for each label $k$. The prediction of the
forest is the label with the highest score $S_k$.

The score $S_k$ mainly depends on the following terms:
\begin{itemize}
		\item $G_j$, the predictive probability at node $j$.
		\item $p_j(x)$ the probability for data point $x$ to generate a split at node $j$.
		\item $P_{NotSperatedYet}(j, x)$ the probability of not having generated a split
				yet as node $j$.
		\item $s_{j,k}(x)$ the score given by a node $j$ to assign label $k$ to data point $x$.
		\item $\eta_j(x)$ the distance to the node's box.
\end{itemize}
The following paragraphs explain how $S_k$ is computed.

$\eta_j(x)$ is computed as follows:
\begin{equation}
		\label{eqt:eta}
		\eta_j(x) = \sum_d (max(x_d - u_{jd}, 0) + max(l_{jd} - x_d, 0)),
\end{equation}
where:
\begin{itemize}
	\item $u_{jd}$ and $l_{jd}$
are respectively the upper and lower bound at node $j$ on feature $d$,
\item $x_d$
is the value of the data point $x$ for feature $d$.
\end{itemize}
\diff{$\eta_j(x)$ depends
on the distance between $x$ and the box of node $j$ defined by $u_j$ and $l_j$ for all
features. If $x$ falls within the box of $j$ then
$\eta_j(x)$ is zero. If $x$ falls outside of
the box of $j$ then $\eta_j(x)$ is positive and it increases with the distance between $x$ and the box containing $j$. $x$ is more likely to be far
from the node box if the node box is small.}

Equation~\ref{eqt:proba_branch} shows how $p_j(x)$ is computed based on
$\eta_j(x)$ and $\Delta_j$, a distance between node $j$ and its parent. We note
that when $x$ falls within the box of $j$, $\eta_j(x)$ is equal to zero, and
thus, $p_j(x)$ is null. \diff{Conversely, when the box size decreases, the
value of $\eta_j(x)$ increases, leading $p_j(x)$
toward 1.
}
\begin{equation}
		\label{eqt:proba_branch}
		p_j(x) = 1 - exp(-\Delta_j \eta_j(x))
\end{equation}

Equation~\ref{eqt:proba_not_branch} defines $P_{NotSperatedYet}(j)$, the
probability of not having branched off before reaching node $j$. It uses
$p_j(x)$, the probability to branch off at node $j$ defined in
Equation~\ref{eqt:proba_branch}, and $ancestors(j)$ which returns the ancestors
of node $j$ starting from the root. \diff{In this
situation, smaller node boxes tend to have a
$p_g(x)$ closer to 1, and thus
$P_{NotSperatedYet}(j, x)$ closer to 0.}
\begin{equation}
		\label{eqt:proba_not_branch}
		P_{NotSperatedYet}(j, x) = \prod_{g \in ancestors(j)} (1 - p_g(x))
\end{equation}

Equation~\ref{eqt:score_node} describes how the score of node $j$ is computed
for label $k$. It uses $p_j(x)$, the probability of branching off at that node,
$P_{NotSperatedYet}(j)$, defined in Equation~\ref{eqt:proba_not_branch}, and
$G_{j,k}$, the predictive probability at node $j$
for label $k$. \diff{If the node box is small,
the weight given to the node predictive
probability $G_{j,k}$ approaches 0 because
$P_{NotSperatedYet}(j, x)$ approaches 0, thus $s_{j,k}(x)$
approaches 0 as well.
}
\begin{equation}
		\label{eqt:score_node}
		s_{j,k}(x) =  
		\left\{
			\begin{array}{ll}
					P_{NotSperatedYet}(j, x) (1 - p_j(x)) G_{j,k} & \mbox{$j =$ leaf}.\\
					P_{NotSperatedYet}(j, x) p_j(x) G_{j,k} & \mbox{otherwise}.
			\end{array}
	\right.
\end{equation}

The score given by a tree for label $k$ and data point $x$, $S_k(x)$, is shown in
Equation~\ref{eqt:score_tree}. $leaf(x)$ returns the leaf node where the data
point $x$ has been sorted to. $path(j)$ returns the list of nodes that lead to
node $j$, starting from the root.
\begin{equation}
		\label{eqt:score_tree}
		S_k(x) = \sum_{j \in path(leaf(x))} s_{j,k}(x)
\end{equation}

We can see from the computation of $S_k$ that the strategies that do not expand
the node boxes (Count Only and Stopped, as well as Partial Update and Ghost
to a certain extent) tend to use the tree root rather than leaves to predict
class labels, even though the root only has a very rough approximation of class
distributions. 
Indeed, in a situation where boxes are maintained small such as with Count Only
strategy, the data point will have a greater chance to fall outside a node box
as well as farther from it. In this case, the probability of branching off
$p_j(x)$, will be higher and thus $P_{NotSperatedYet}(j)$ will be lower as we go
deeper in the tree.

Therefore, when computing $S_k(x)$, the node score $s_{j,k}$ will become
smaller as we get closer to a leaf because $s_{j,k}$ is the product of the
node prediction $G_{j,k}$ weighted by $P_{NotSperatedYet}(j, x)$. In that
situation, most of the score comes from nodes closer to the root, but these
nodes have a poor approximation of the feature space.

For nodes closer to the leaf, analysis shows that their weight in the tree
prediction increases as box sizes increase. In general, with a smooth label
distribution, we think that Extend Node would exhibit better performance for
this reason. We also expect these nodes to become irrelevant in a concept drift
situation. In this scenario, using nodes closer to the root may be more
relevant.

The following empirical study intends to compare the different approaches and
determine if it is better to increase the impurity of the node by forcing the
box to extend (Extend Node strategy) and include data points that should have
branched off than keeping the box small and having a finer-grain partition of
the space (Count Only, Stopped, Partial Update, and Ghost).

\vspace{1cm}
\subsection{Datasets}
\diff{We used 49 datasets to evaluate our proposed
methods: 42 synthetic datasets to mimic real-world
situations and to make comparisons with and
without concept drifts, and 7 real
datasets.}

\subsubsection{\banosdataset}
The \banosdataset~dataset~\cite{Banos_2014}\footnote{available
\href{https://archive.ics.uci.edu/ml/datasets/REALDISP+Activity+Recognition+Dataset\#:\~:text=The\%20REALDISP\%20(REAListic\%20sensor\%20DISPlacement,\%2Dplacement\%20and\%20induced\%2Ddisplacement}{here}}
is a human activity dataset with 17 participants and 9~sensors per participant.
Each sensor samples a 3D acceleration, gyroscope, and magnetic field, as well as
the orientation in a quaternion format, producing a total of 13 values.  Sensors
are sampled at 50~Hz, and each sample is associated with one of 33 activities.
In addition to the 33 activities, an extra activity labeled 0 indicates no
specific activity.

We pre-processed the \banosdataset dataset as in \cite{Banos_2014}, using non-overlapping windows of one
second (50~samples), and using only the 6 axes (acceleration and gyroscope) of
the right forearm sensor. We computed the average and standard deviation over
the window as features for each axis. We assigned the most frequent label to the
window. The resulting data points were shuffled uniformly.

In addition, we constructed another dataset from \banosdataset, in which we
simulated a concept drift by shifting the activity labels in the second half of
the data stream. 

\subsubsection{\recofitdataset}
The \recofitdataset dataset~\cite{recofit, recofit_data} is a human activity dataset
containing 94 participants.
Similar to the \banosdataset dataset, the activity labeled 0 indicates no
specific activity. Since many of these activities were similar, we merged  some
of them together based on the table in~\cite{cross_subject_validation}. 

We pre-processed the dataset similarly to the \banosdataset dataset, using
non-overlapping windows of one second, and only using 6 axes (acceleration and
gyroscope) from one sensor. From these 6 axes, we used the average and the
standard deviation over the window as features. We assigned the most frequent
label to the window.

\diff{
\subsubsection{PAMAP2}
The PAMAP2 dataset~\cite{pamap2} is a human
activity recognition dataset that comprises data
collected from 9 participants. Similar to the
\banosdataset dataset, the activity labeled 0
indicates no specific activity. While the original paper mentions that 18
activities were performed, only 12 activities are
available in the dataset.

Similarly to \banosdataset dataset, we
pre-processed the dataset using non-overlapping
windows of one second and using data from six axes
obtained from a single sensor placed on the chest.
Features were extracted using the average and
standard deviation for each axes over each window,
and the most frequent activity label was assigned
to the window. 
}
\diff{
\subsubsection{HARTH}
The HARTH dataset~\cite{harth} is a human activity
recognition dataset containing data from 22
participants who performed 12 activities. The
dataset was pre-processed similarly to the
\banosdataset dataset, using non-overlapping
windows of one second. However, the wearable
sensor placed on the lower back of the
participants only recorded acceleration data from
three axes. Features were extracted using the
average and standard deviation over each window,
and the most frequent activity label was assigned
to the window.
}
\diff{
\subsubsection{HAR70+}
The HAR70+ dataset~\cite{har70} is a Human Activity
Recognition dataset containing data from 18
participants between the ages of 70 and 95, who
performed 8 activities. During data recording,
some participants used walking aids. The dataset was
pre-processed similarly to the \banosdataset dataset,
using non-overlapping windows of one second. The
wearable sensor placed on the thigh only recorded
acceleration data from three axes. Features were
extracted using the average and standard deviation
over each window, and the most frequent activity
label was assigned to the window.
}

\subsubsection{Covtype}
The \covtypedataset dataset\footnote{available
\href{https://archive.ics.uci.edu/ml/datasets/covertype}{here}} is a tree dataset. Each data
point is a tree described by 54 features including ten quantitative variables
and 44 binary variables. The 581,012 data points are labeled with one of the seven forest cover
types and these labels are highly imbalanced. In particular, two labels represent
85\% of the dataset.

\subsubsection{Synthetic Datasets}
\diff{We generated 42 validation datasets using
RandomRBF, SEA, SINE, and Hyperplane generators
from Massive Online Analysis (MOA)~\cite{moa}. We
also randomized their parameters by changing the
seed and randomizing the number of centroids for
RandomRBF, the function and noise level for SEA,
the function for SINE, and the noise level and the
number of drifting attributes for Hyperplane. The
noise level was selected between 0\%, 4\%, and
8\%. The number of centroids was selected between
34 and 200. Half of the Hyperplane datasets have a
concept drift of 0.001 and their number of
drifting attributes was selected between 1 and 7.
We generated 20,000 data points for each of these
synthetic datasets. MOA commands are available
}
\href{https://github.com/big-data-lab-team/benchmark-har-data-stream/blob/956e81f446a531111c1680a1abb96d6117a19a87/Makefile#L200}{here}.

\subsection{Evaluation Metric}
We evaluated our methods using a prequential fading macro F1-score. We used a
prequential metric because data stream models cannot be evaluated with the
traditional testing/training sets since the model continuously
learns from a stream of data points~\cite{issues_learning_from_stream}. We focused on the F1 score because
most datasets are imbalanced. We used the prequential version of the F1 score to
evaluate classification on data stream. We used a fading factor to minimize the
impact of old data points, especially data points at the beginning or data
points saw before a drift occur. To obtain this fading F1 score, we multiplied
the confusion matrix with the fading factor before incrementing the cell in the
confusion matrix.

\begin{figure*}[!t]
\centering
	\includegraphics[width=0.85\textwidth]{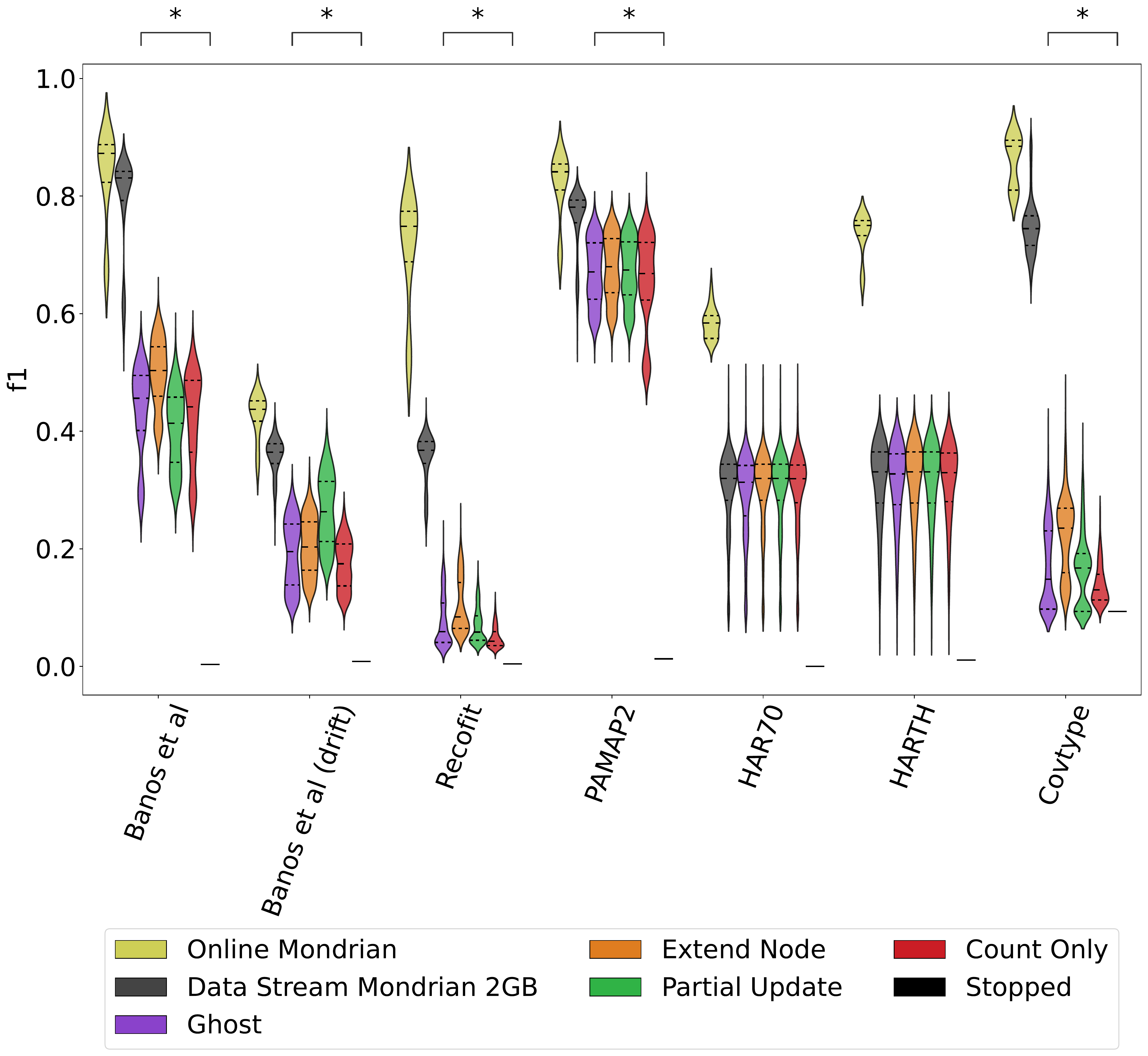}
    \caption{\diff{Comparison of out-of-memory strategies for the
    real datasets (n=200 repetitions). The star annotation indicates
    statistically significant differences
    (Bonferroni corrected) between
    all pairs of proposed out-of-memory strategies
    (Ghost, Extend Node, Partial Update, Count
    Only, and Stopped).}}
	\label{fig:f1_xp1_real}
\end{figure*}

\section{Results}
Our experiments evaluated the out-of-memory and adaptation strategies
presented previously. We evaluated both sets of methods separately. Unless
specified otherwise, we allocated 600~KB of memory for the forest, which
allowed for 940 to 1600 nodes in the forest depending on the number of
features, labels, and trees. As a comparison, the
Raspberry Pi Pico has 256~KB of memory and the Arduino has
between 2~KB (Uno) and 8~KB (MEGA 2560).

\subsection{Baselines}
\diff{Figure~\ref{fig:f1_xp1_real} shows the F1 score
distributions obtained at the end of each dataset. The experiment includes the
methods described in Section~\ref{sec:method-xp1} --- executed wih a memory
limit of 0.6~MB --- as well as the online \mondrianforest and the Data Stream
\mondrianforest with 2~GB of memory for reference. The limit of 2~GB is reached
for the \covtypedataset and \recofitdataset datasets, and the Data Stream Mondrian 2~GB
method applies the Extend Node method in that situation. The Online Mondrian
is the Python implementation available on
GitHub~\cite{mondrian_implementation_1} that we
modified to output the offline F1
score.}

The Online Mondrian reaches state-of-the-art performance in
the real datasets~\cite{StreamDM-CPP,
fast_and_slow,kappa_updated_ensemble, cross_subject_validation}.
\diff{The Data Stream Mondrian~2~GB performance is significativly lower
than the Online Mondrian. These differences are
explained by two factors:}
the Online Mondrian is evaluated with a holdout
set randomly selected from the dataset whereas the
Data Stream Mondrian is evaluated with a
prequential method (as is commonly done in data
streams); the Online Mondrian can access to
previous data points while the Data Stream
Mondrian cannot. The Online Mondrian is given here
as a reference for comparison, however, it should
not be directly compared to the Data Stream
Mondrian as these methods operate in different
contexts.

The Stopped method, the default reference for
evaluation under memory constraints, has by far
the lowest F1 score, which demonstrates the
usefulness of our out-of-memory strategies. The
impact of memory limitation is clear and can be
seen by the substantial performance edge of Data
Stream Mondrian 2~GB over the other methods.

\subsection{Out-of-memory strategies}
\label{sec:xp1_result}
\begin{figure*}[!t]
\centering
	\includegraphics[width=0.85\textwidth]{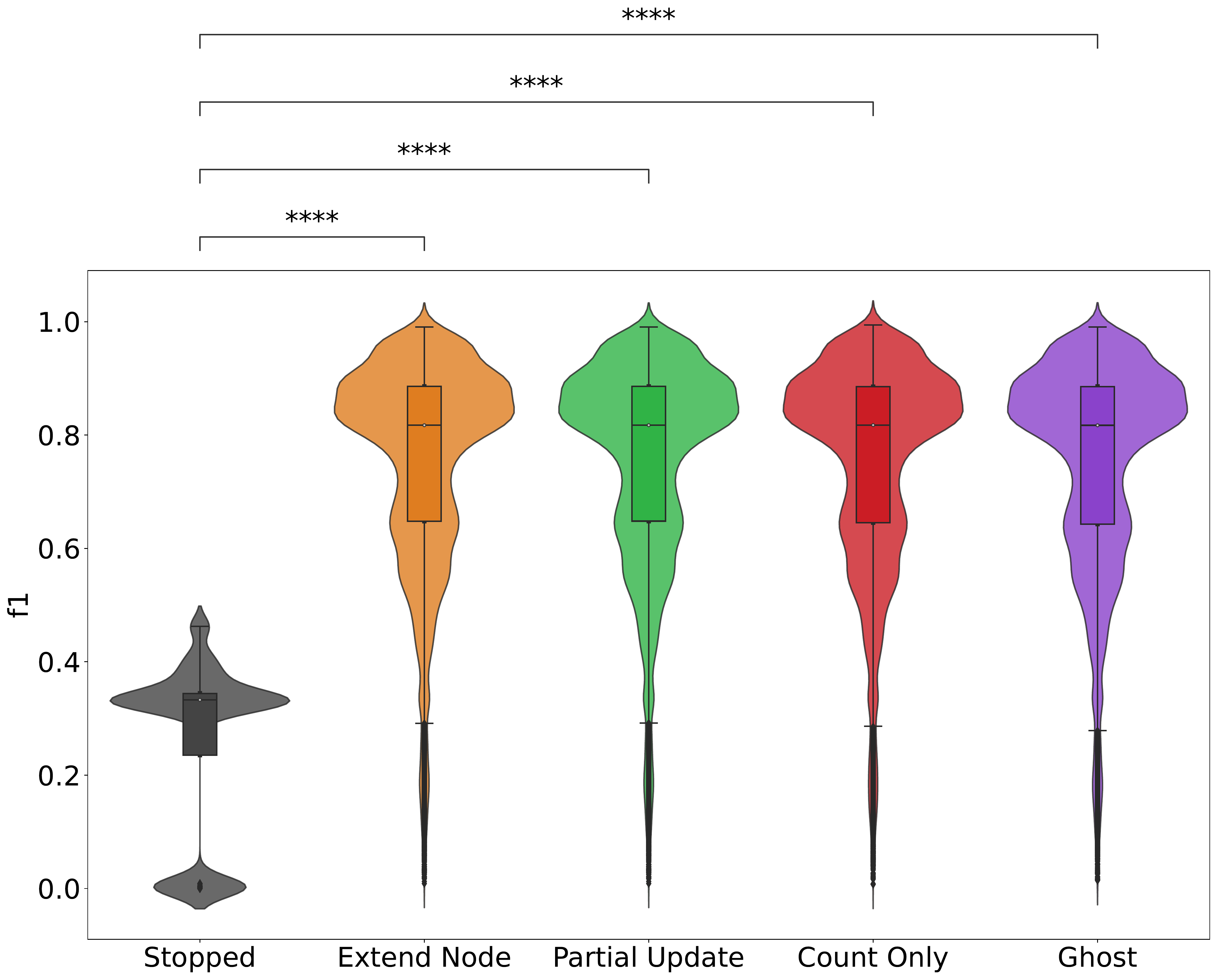}
		\caption{\diff{Comparison of out-of-memory strategies for the 40
		synthetic datasets generated with MOA (n=200 repetitions).}}
	\label{fig:f1_xp1_violin_synthetic}
\end{figure*}

\diff{
The star annotation in
Figure~\ref{fig:f1_xp1_real} highlights
statistically significant differences among the
five proposed out-of-memory strategies (Ghost,
Extend Node, Partial Update, Count Only, and
Stopped). No significant differences are observed
between the methods on the HAR70 and HARTH
datasets, except for Stopped, which performs
significantly lower than the others.

In the other datasets, the Extend Node method
consistently stands above the other ones except
for the \banosdataset with drift dataset where the
Partial Update method achieves the best F1 score.
This is due to a faster recovery of the nodes'
counters after the drift. The Extend Node makes the
best of two factors. First, it does not drop any
data points compared to Partial Update or Count
Only. Second, since it extends the node boxes, future
data points fall within the box and receive the
prediction of the corresponding leaf, whereas the
Count Only and the Ghost methods soften the leaf
prediction depending on the data point distance
with the box.

Figure~\ref{fig:f1_xp1_violin_synthetic} depicts
the F1 score distributions for 40 synthetic
datasets generated with MOA. All our proposed out-of-memory strategies achieve
similar F1 scores --- clearly better than the
default Stopped method. Additionally, no
significant difference is observed between Extend
Node, Partial Update, Ghost, and Count Only across
these datasets. However, it is noteworthy that the
Stopped method exhibits significantly lower
performance compared to the other strategies.
}

\diff{The difference in F1 score between the Stopped Mondrian and the other methods is
reported in Table~\ref{tab:delta-f1-xp1} for the
real datasets}. The difference is computed for all
numbers of trees and we report the minimum, the mean, and the maximum. On average,
 the Extend Node has the best improvement over
Stopped Mondrian with an average improvement in F1-score of 0.31,
a minimum of 0.1, and a maximum of 0.66.

\begin{table}
\begin{center}
\resizebox{\columnwidth}{!}{
\begin{tabular}{ | r | l | c c c |}
\hline
            &         &     & $\Delta$ F1 & \\
Dataset & Method Name & Min & Mean     & Max\\
\hline
Banos et al & Count Only     & .23 & .42 & .56\\
            & Extend Node    & \textbf{.35} & \textbf{.49} & \textbf{.63}\\
            & Ghost          & .24 & .43 & .56\\
            & Partial Update & .26 & .4 & .57\\
\hline
Banos et al (drift) & Count Only     & .07 & .17 & .27\\
                    & Extend Node    & .09 & .2 & .32\\
                    & Ghost          & .08 & .18 & .31\\
                    & Partial Update & \textbf{0.13} & \textbf{0.26} & \textbf{0.4}\\
\hline
Covtype & Count Only     & .0 & .05 & .18\\
        & Extend Node    & .0 & \textbf{.13} & \textbf{.37}\\
        & Ghost          & .0 & .07 & .31\\
        & Partial Update & .0 & .07 & .29\\
\hline
HAR70 & Count Only     & \textbf{.09} & \textbf{.3} & \textbf{.48}\\
      & Extend Node    & \textbf{.09} & \textbf{.3} & \textbf{.48}\\
      & Ghost          & \textbf{.09} & .29         & \textbf{.48}\\
      & Partial Update & \textbf{.09} & \textbf{.3} & \textbf{.48}\\
\hline
HARTH & Count Only     & \textbf{.04} & \textbf{.3} & \textbf{.42}\\
      & Extend Node    & \textbf{.04} & \textbf{.3} & \textbf{.42}\\
      & Ghost          & \textbf{.04} & \textbf{.3} & .41\\
      & Partial Update & \textbf{.04} & \textbf{.3} & \textbf{.42}\\
\hline
PAMAP2 & Count Only     & .47 & .64 & \textbf{.79}\\
       & Extend Node    & \textbf{.53} & \textbf{.66} & .77\\
       & Ghost          & \textbf{.53} & \textbf{.66} & .77\\
       & Partial Update & \textbf{.53} & \textbf{.66} & .77\\
\hline
Recofit & Count Only     & .02 & .04 & .11\\
        & Extend Node    & \textbf{.04} & \textbf{.1}  & \textbf{.25}\\
        & Ghost          & .02 & .07 & .22\\
        & Partial Update & .03 & .06 & .16\\
\hline
\end{tabular}
}
\end{center}
\caption{$\Delta$F1 score compared to Stopped Mondrian. Minimum, maximum, and average scores are computed across all tree numbers.}
\label{tab:delta-f1-xp1}
\end{table}

From these results, we conclude that the Extend
Node should be the default strategy to follow when
the \mondrianforest reaches the memory limit.

\subsection{Concept Drift Adaptation for Mondrian Forest under Memory Constraint}
\begin{figure*}[!t]
	\includegraphics[width=\textwidth]{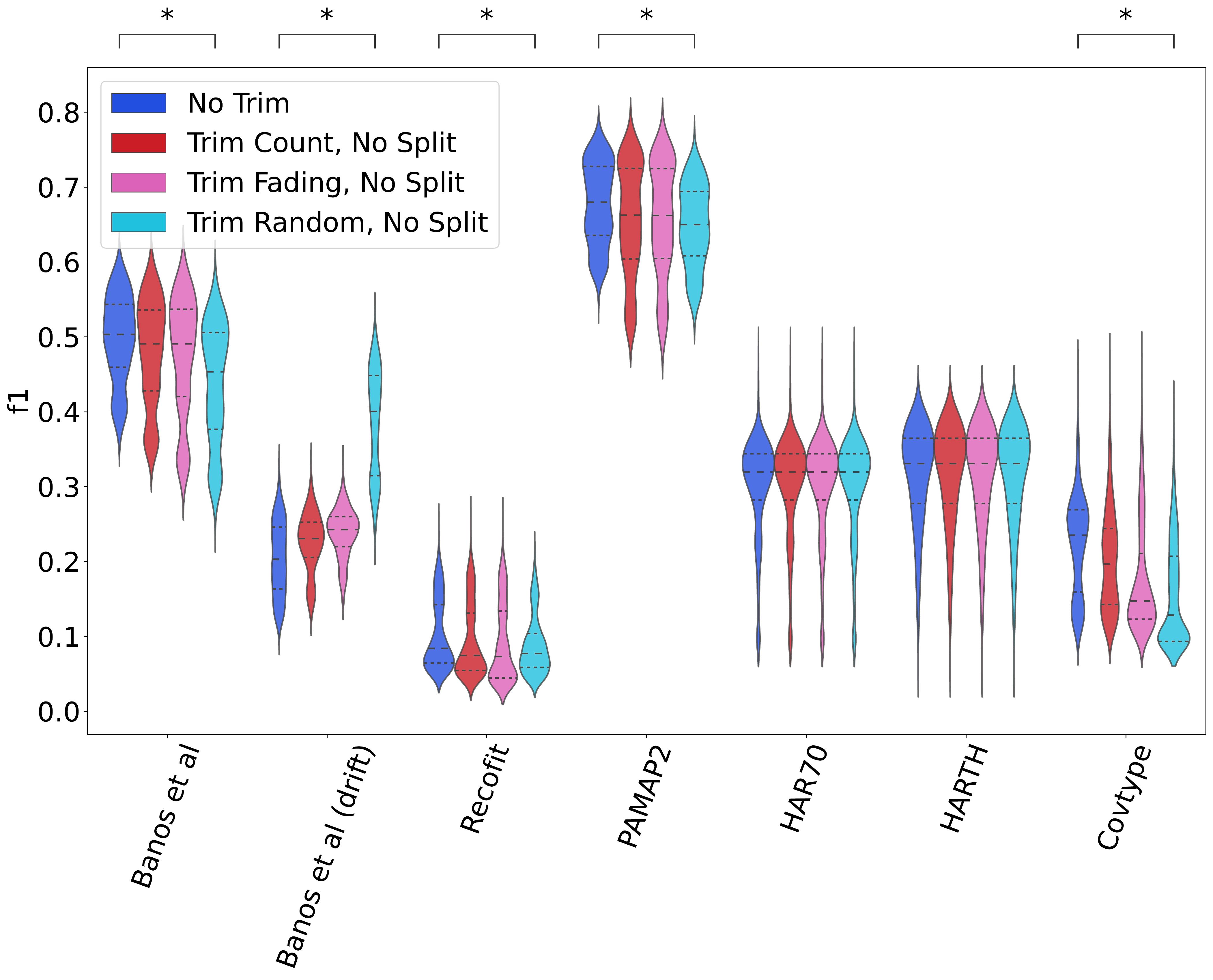}
		\caption{\diff{Comparison of node trimming strategies for the real datasets (n=200 repetitions). The star annotation indicates
		statistically significant differences (Bonferroni corrected).}}
	\label{fig:f1-xp2-1}
\end{figure*}

\diff{Figure~\ref{fig:f1-xp2-1} shows the F1 score for the three proposed
leaf trimming strategies. All trimming strategies were evaluated with the Extend Node
out-of-memory strategy as it outperforms the other out-of-memory strategies per
our previous experiment. In the case of
\banosdataset with drift, the random trimming method outperforms the other ones
whereas for the stable datasets (\banosdataset, \covtypedataset,
\recofitdataset), no trimming appears to be the best strategy.
}

\begin{figure*}[!t]
  \includegraphics[width=\textwidth]{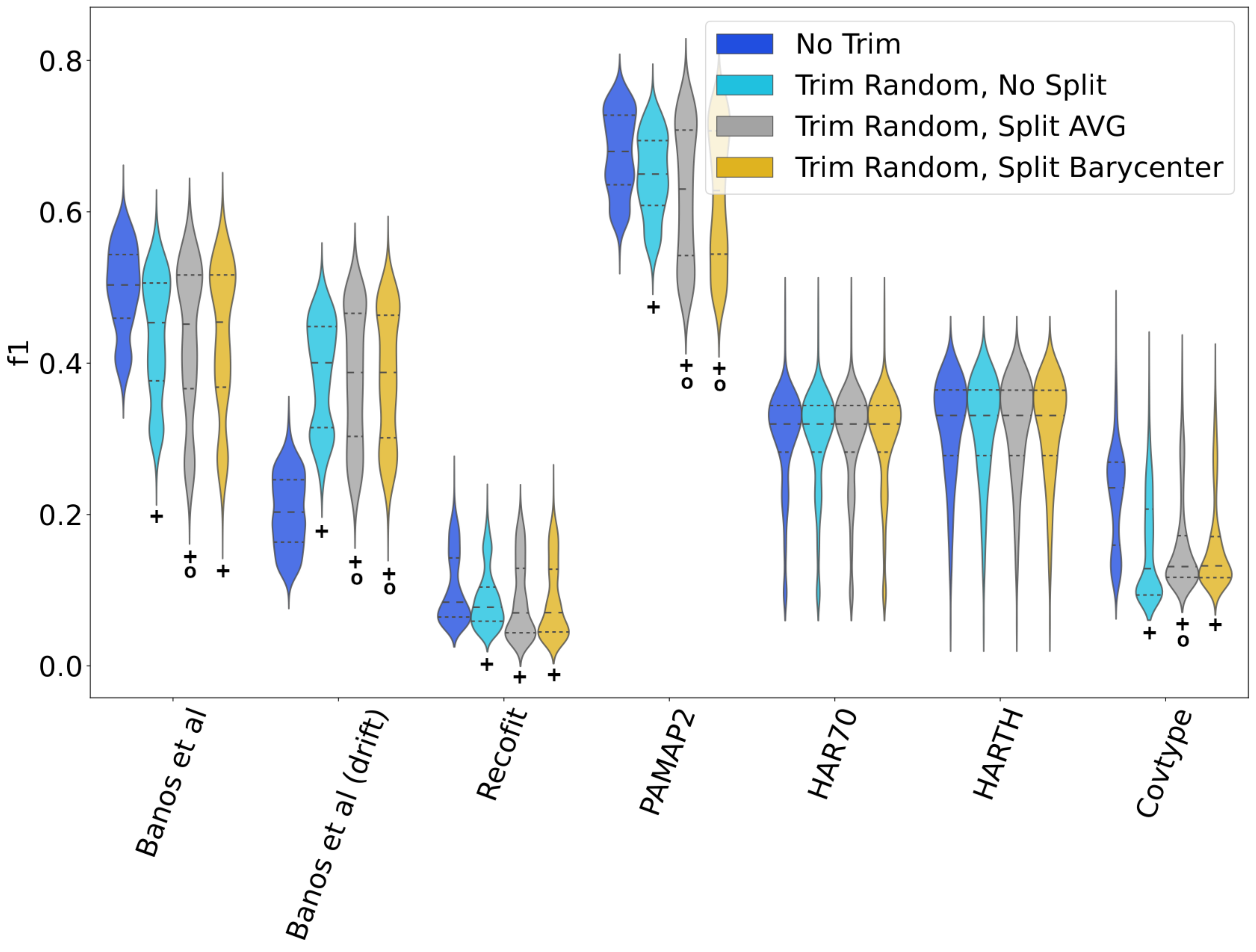}
  \caption{\diff{Comparison of node splitting strategies (n=200 repetitions). The cross below the
  violin indicates a statistical significant
  difference between, the method and No Trim,
  whereas the circle indicates a difference with No
  Split.}}
  \label{fig:f1-xp2-3}
\end{figure*}

\diff{Figure~\ref{fig:f1-xp2-3} shows the F1 score for
the three splitting methods (no split, Split AVG,
and Split Barycenter) combined with random
trimming and extend node as the out-of-memory
strategy. With the drift dataset, the three
methods perform better than not trimming. Finally,
with \banosdataset, \recofitdataset, and
\covtypedataset datasets, not trimming is
generally better than randomly trimming.}

\begin{figure*}[!t]
	\includegraphics[width=\textwidth]{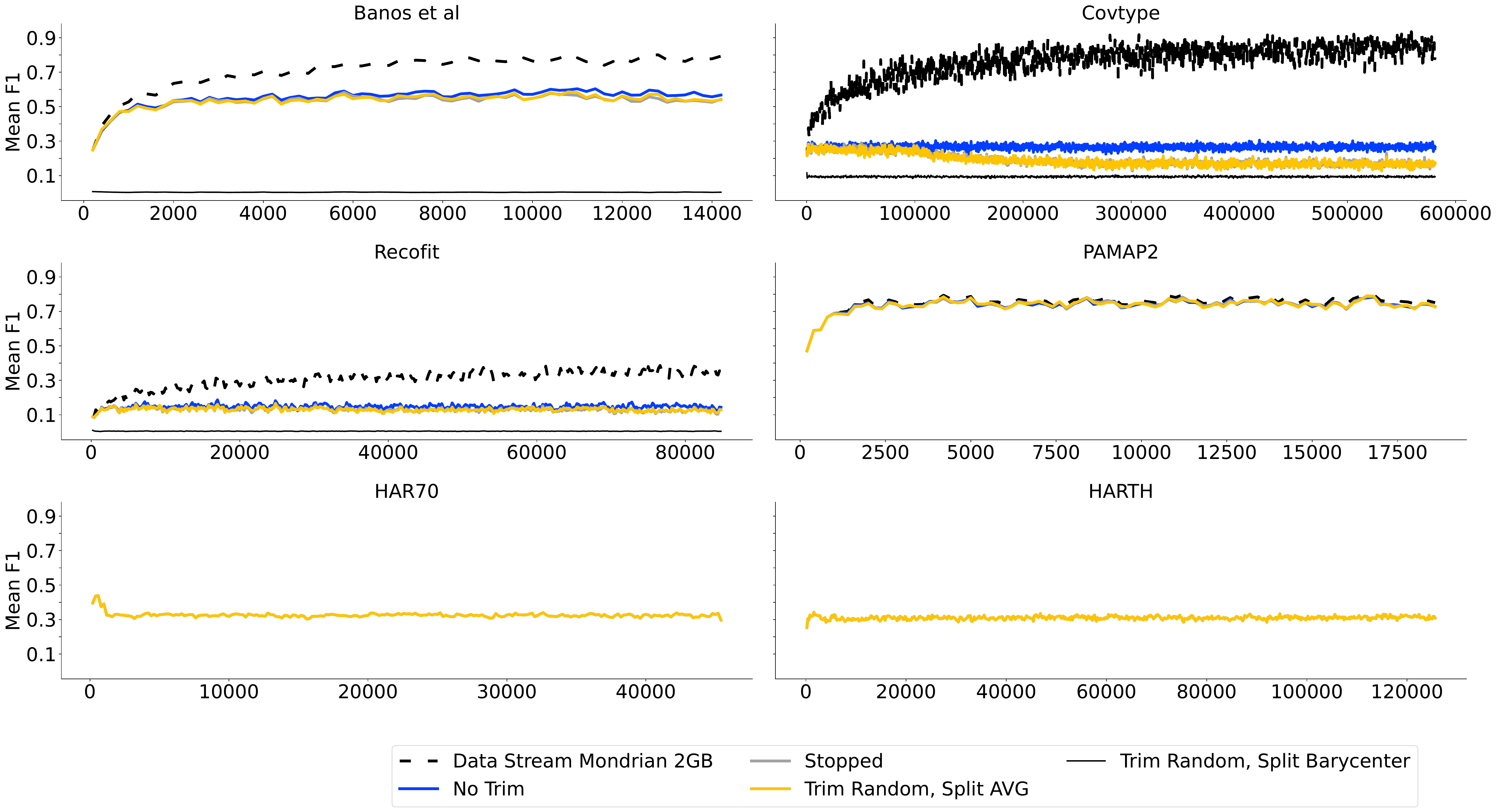}
    \caption{\diff{ The Evolution of F1 score over
    time for a 5-Tree \mondrianforest with stable
    datasets (n=10 repetitions).}}
	\label{fig:f1-xp2-5-stable}
\end{figure*}

\begin{figure*}[!t]
	\includegraphics[width=\textwidth]{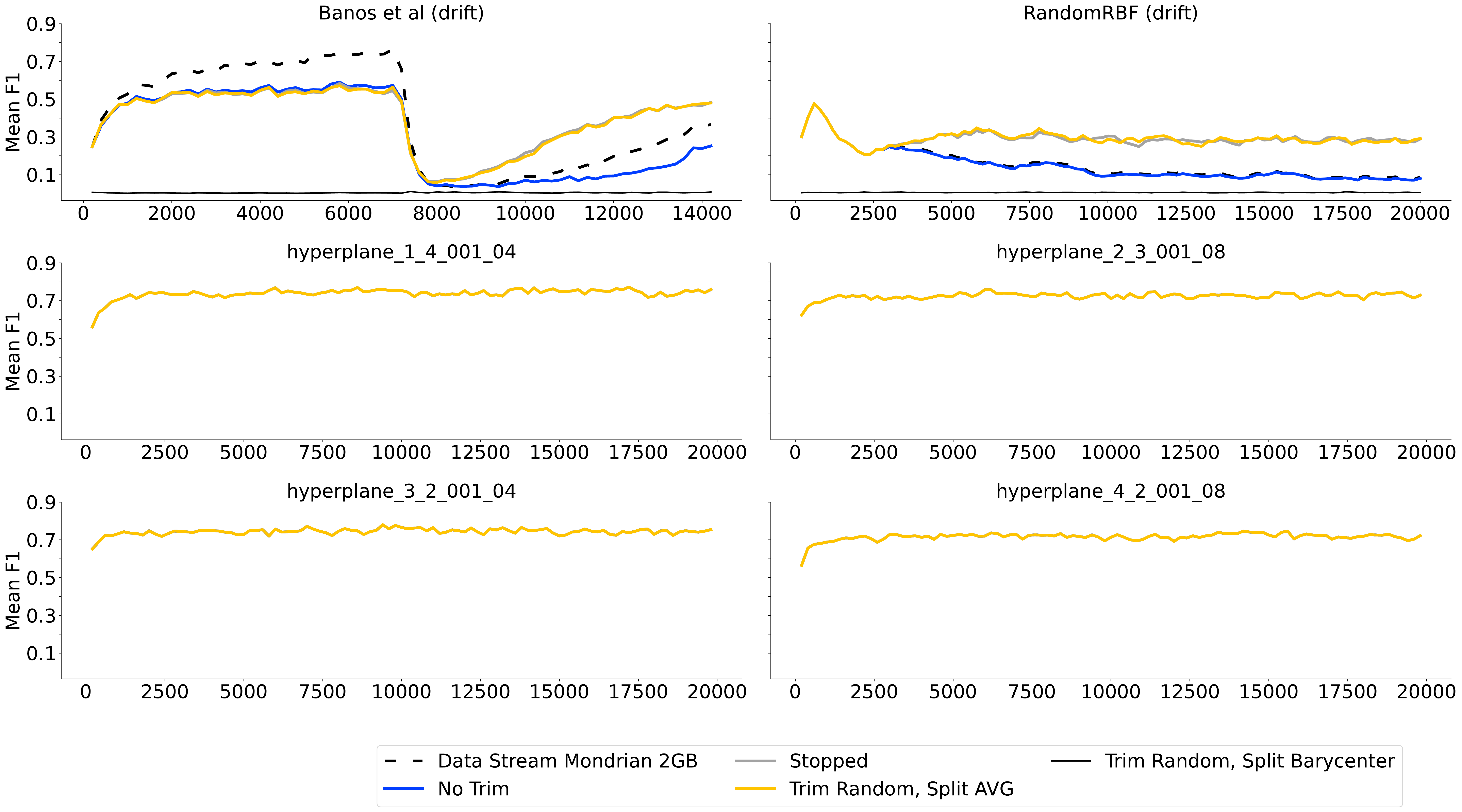}
    \caption{\diff{ The evolution of F1 score over
    time for a 5-Tree \mondrianforest with drift
    datasets (n=10 repetitions).}}
	\label{fig:f1-xp2-5-drift}
\end{figure*}

\diff{
Figure~\ref{fig:f1-xp2-5-stable} illustrates the
evolution of the F1 score over time using 5-tree
\mondrianforests with real datasets. We observe
that the Mondrian~2GB consistently achieves a
higher plateau in three of the datasets (Banos et
al, Covertype, and Recofit). Additionally, the two
Trim Random methods exhibit a decline in F1 score
on two datasets (\banosdataset and Covtype).
Interestingly, all methods yield similar results
with the PAMAP2, HAR70, and HARTH datasets,
suggesting that the features may not adequately
capture the label information.

In Figure~\ref{fig:f1-xp2-5-drift}, we present the
F1 score's evolution over time using 5-tree
Mondrian forests with drift datasets. Notably, the
\banosdataset (drift) dataset experiences a sudden
drift that impacts all methods, but they manage to
recover, with the Trim Random methods exhibiting a
faster recovery. For the RandomRBF dataset, all
methods show a simultaneous drop in performance,
but only the Trim Random methods demonstrate
recovery from the gradual drift. Additionally, the
figure includes the results for the Hyperplane
datasets, where all methods perform similarly.
}


Table~\ref{tab:delta-f1-xp2} summarizes the delta of F1 scores of the trimming
methods compared to the Stopped method. From these results, we conclude that
randomly trimming allows the \mondrianforest to adapt to concept drift. We also
conclude that using the split methods (Split AVG and Split Barycenter) should be
the default method to grow the \mondriantrees after trimming as they exhibit a
better F1 score than not splitting.

\begin{table}
\begin{center}
\resizebox{\columnwidth}{!}{
\begin{tabular}{ | r | l | c c c |}
\hline
            &         &     & $\Delta$ F1 & \\
Dataset & Method Name & Min & Mean     & Max\\
\hline
\banosdataset & Trim Count, No Split & .32 & .48 & \textbf{.62}\\
            & Trim Count, Split AVG & .33 & \textbf{.49} & \textbf{.62}\\
            & Trim Count, Split Barycenter & \textbf{.34} & \textbf{.49} & \textbf{.62}\\
            & Trim Fading, No Split & .29 & .47 & .61\\
            & Trim Fading, Split AVG & .31 & .48 & .61\\
            & Trim Fading, Split Barycenter & .3 & .48 & \textbf{.62}\\
            & Trim Random, No Split & .25 & .43 & .59\\
            & Trim Random, Split AVG & .21 & .43 & .59\\
            & Trim Random, Split Barycenter & .19 & .43 & .6\\
\hline
\banosdataset (drift) & Trim Count, No Split & .11 & .22 & .33\\
                    & Trim Count, Split AVG & .15 & .28 & .39\\
                    & Trim Count, Split Barycenter & .14 & .28 & .38\\
                    & Trim Fading, No Split & .13 & .23 & .33\\
                    & Trim Fading, Split AVG & .2 & .38 & .53\\
                    & Trim Fading, Split Barycenter & .18 & \textbf{.39} & .52\\
                    & Trim Random, No Split & \textbf{.22} & .38 & .52\\
                    & Trim Random, Split AVG & .19 & .37 & .53\\
                    & Trim Random, Split Barycenter & .17 & .37 & \textbf{.54}\\
\hline
Covtype & Trim Count, No Split & .0 & .11 & .38\\
        & Trim Count, Split AVG & .0 & \textbf{.12} & .37\\
        & Trim Count, Split Barycenter & .0 & \textbf{.12} & .37\\
        & Trim Fading, No Split & .0 & .08 & .38\\
        & Trim Fading, Split AVG & .0 & .\textbf{12} & \textbf{.39}\\
        & Trim Fading, Split Barycenter & .0 & \textbf{.12} & .37\\
        & Trim Random, No Split & .0 & .06 & .31\\
        & Trim Random, Split AVG & \textbf{.01} & .07 & .31\\
        & Trim Random, Split Barycenter & .0 & .07 & .3\\
\hline
HAR70 & Trim Count, No Split & \textbf{.09} & \textbf{.3} & \textbf{.48}\\
      & Trim Count, Split AVG & \textbf{.09} & \textbf{.3} & \textbf{.48}\\
      & Trim Count, Split Barycenter & \textbf{.09} & \textbf{.3} & \textbf{.48}\\
      & Trim Fading, No Split & \textbf{.09} & \textbf{.3} & \textbf{.48}\\
      & Trim Fading, Split AVG & \textbf{.09} & \textbf{.3} & \textbf{.48}\\
      & Trim Fading, Split Barycenter & \textbf{.09} & \textbf{.3} & \textbf{.48}\\
      & Trim Random, No Split & \textbf{.09} & \textbf{.3} & \textbf{.48}\\
      & Trim Random, Split AVG & \textbf{.09} & \textbf{.3} & \textbf{.48}\\
      & Trim Random, Split Barycenter & \textbf{.09} & \textbf{.3} & \textbf{.48}\\
\hline
HARTH & Trim Count, No Split & \textbf{.04} & \textbf{.3} & \textbf{.42}\\
      & Trim Count, Split AVG & \textbf{.04} & \textbf{.3} & \textbf{.42}\\
      & Trim Count, Split Barycenter & \textbf{.04} & \textbf{.3} & \textbf{.42}\\
      & Trim Fading, No Split & \textbf{.04} & \textbf{.3} & \textbf{.42}\\
      & Trim Fading, Split AVG & \textbf{.04} & \textbf{.3} & \textbf{.42}\\
      & Trim Fading, Split Barycenter & \textbf{.04} & \textbf{.3} & \textbf{.42}\\
      & Trim Random, No Split & \textbf{.04} & \textbf{.3} & \textbf{.42}\\
      & Trim Random, Split AVG & \textbf{.04} & \textbf{.3} & \textbf{.42}\\
      & Trim Random, Split Barycenter & \textbf{.04} & \textbf{.3} & \textbf{.42}\\
\hline
PAMAP2 & Trim Count, No Split & .48 & .64 & .77\\
       & Trim Count, Split AVG & .48 & \textbf{.65} & .78\\
       & Trim Count, Split Barycenter & .48 & .64 & .78\\
       & Trim Fading, No Split & .47 & .64 & .77\\
       & Trim Fading, Split AVG & .47 & .63 & \textbf{.79}\\
       & Trim Fading, Split Barycenter & .47 & .63 & .78\\
       & Trim Random, No Split & \textbf{.5} & .64 & .76\\
       & Trim Random, Split AVG & .44 & .61 & .77\\
       & Trim Random, Split Barycenter & .44 & .61 & .77\\
\hline
\recofitdataset & Trim Count, No Split & .03 & .09 & \textbf{.26}\\
        & Trim Count, Split AVG & \textbf{.04} & \textbf{.1} & .25\\
        & Trim Count, Split Barycenter & \textbf{.04} & \textbf{.1} & .25\\
        & Trim Fading, No Split & .03 & .09 & \textbf{.26}\\
        & Trim Fading, Split AVG & .03 & \textbf{.1} & .24\\
        & Trim Fading, Split Barycenter & .03 & \textbf{.1} & .24\\
        & Trim Random, No Split & .03 & .08 & .22\\
        & Trim Random, Split AVG & .02 & .08 & .21\\
        & Trim Random, Split Barycenter & .02 & .08 & .24\\
\hline
\end{tabular}
}
\end{center}
\caption{$\Delta$F1 score compared to Stopped Mondrian for the
trimming methods. Minimum, maximum, and average scores are computed across all tree numbers.}
\label{tab:delta-f1-xp2}
\end{table}

\subsection{Impact of the Memory Limit}
We note in Figures~\ref{fig:f1_xp1_real},~\ref{fig:f1-xp2-1}, and~\ref{fig:f1-xp2-3}, that the F1-scores tend to be low even though the Data Stream Mondrian 2~GB reaches state-of-the-art F1-scores.
This implies that reducing the memory from 2~GB to 600~KB has a strong impact on
the performance. We also note that this impact varies between datasets. For the
RandomRBF stable dataset, the methods are closer to the Data Stream Mondrian
2~GB compared to the \covtypedataset dataset where there is a more important difference.

\begin{figure*}[!t]
\centering
	\includegraphics[width=\textwidth]{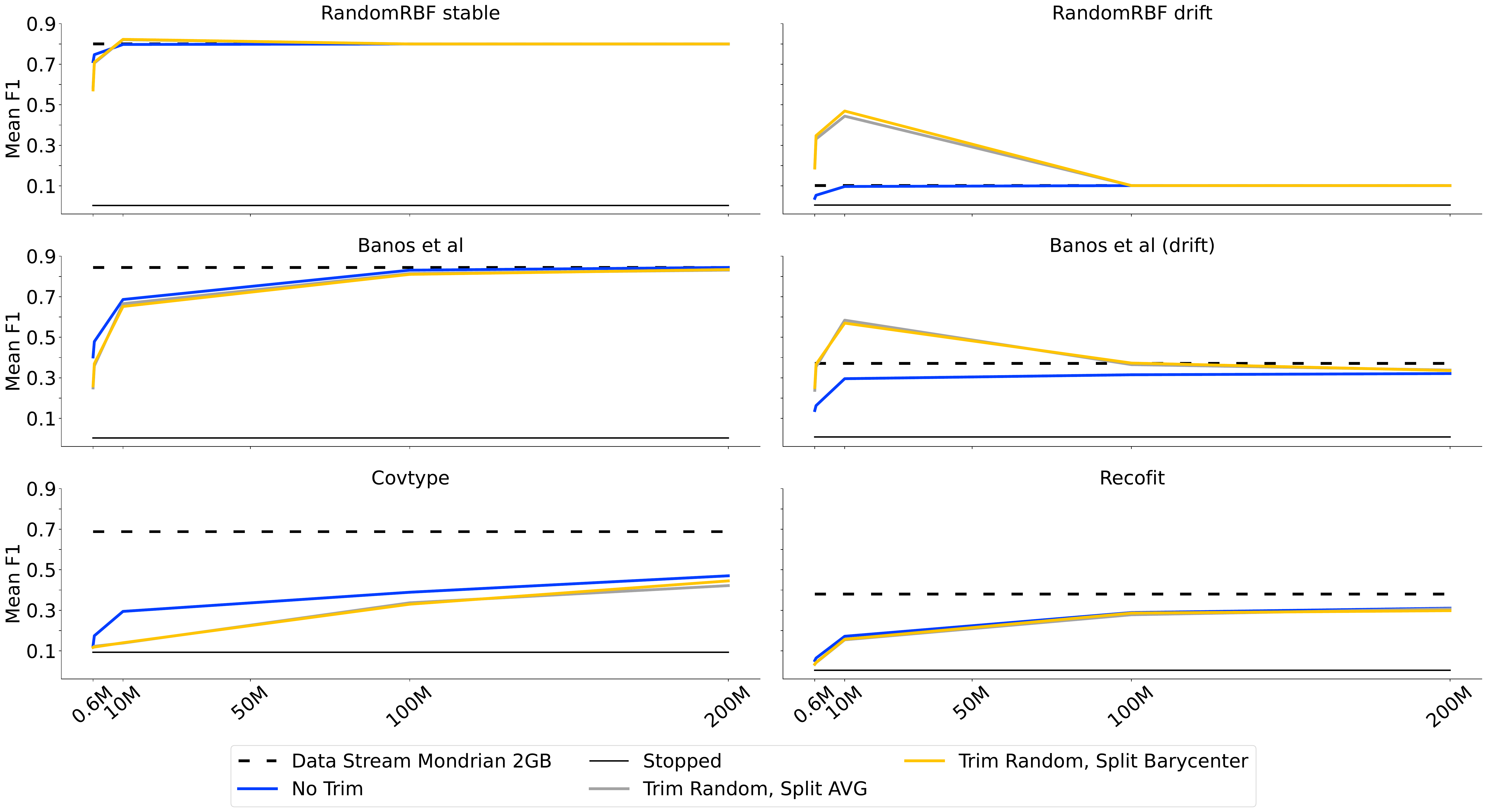}
	\caption{Evaluation of the memory impact on the top out-of-memory strategy and the
		top trimming methods described in Section~\ref{sec:method-xp1} and
		Section~\ref{sec:method-xp2} (n=10 repetitions). The results are shown for 50 trees.}
	\label{fig:f1_xp2-4}
\end{figure*}

Figure~\ref{fig:f1_xp2-4} shows the evolution of the memory limit impact on the
F1-score for a \mondrianforest with 50 trees. We selected 50 trees as it is the
number of trees that benefit the most from more memory. Indeed, more memory
means they are less likely to underfit and thus, have better performance than
fewer trees. Similar trends are observed with fewer trees. The dashed black line
indicates the F1-score reached by Data Stream Mondrian 2~GB.

We note that, except for the Stopped method, F1-scores increase with the amount
of available memory. This is explained by the fact that trees can grow more
nodes and therefore describe a finer-grained partition of the feature space.

We observe that the sharpest improvement for the No Trim method happens between
600~KB and 10~MB, after which the F1-score increase slows down and plateaus
toward the Data Stream Mondrian 2~GB F1-score.

The trimming methods with split exhibit two behaviors. For stable datasets, they
follow the trend of No Trim. For the drift datasets, the trimming methods
improve beyond the Data Stream Mondrian 2~GB up to the 10~MB limit, after which
the F1-score decreases down to the Data Stream Mondrian 2~GB.

This behavior is explained by the trimming algorithm that trims based on the
tree count rather than the tree size. When too much memory is allocated the size
of the trees becomes too big for the trimming pace. Therefore, the old concept
is not trimmed out fast enough, which explains the decrease in F1-score beyond
the 10~MB mark for the trimming methods on the drift datasets.

\diff{
Table~\ref{tab:runtime-xp2} presents the total
runtime of Trim Random with Split Barycenter and No Trim on
the \banosdataset dataset. The runtime is
influenced by two main factors: the number of
trees and the amount of available memory. It’s
worth noting that the Extend Node method, due to
its predictable memory access patterns, can be
optimized to run faster than a \mondrianforest
with trimming enabled. For instance, a
\mondrianforest with 50 trees and 2GB of memory
takes only 9 seconds to run with Extend Node,
while it requires 442 seconds with trimming
enabled. 
}

\begin{table}
\begin{center}
\resizebox{\columnwidth}{!}{
\begin{tabular}{ |c|r c | c c c c c |}
\hline
						    &   & Memory & 0.6MB & 1MB & 10MB & 100MB & 200MB \\
                &Tree Count  &        &       &     &      &       &\\
\hline
                & 1 &       &  0.59  &  0.87   &  2.77     & 2.78 & 2.89  \\
                & 5 &       &  0.78  &  1.03   &  20.35    & 45   & 44    \\
Trim Random     & 10&       &  0.91  &  1.15   &  20.12    & 93   & 91    \\
Split Barycenter& 20&       &  0.97  &  1.34   &  20.89    & 185  & 185   \\
                & 30&       &  1.09  &  1.55   &  22.27    & 230  & 277   \\
                & 50&       &  1.25  &  1.82   &  23.97    & 238  & 442   \\
\hline
                & 1 &       &  0.65  &  0.99   &  5.47     & 5.68 & 5.68  \\
                & 5 &       &  0.77  &  1.46   &  25       & 49   & 51\\
No Trim         & 10&       &  0.90  &  1.53   &  24.75    & 102  & 107\\
                & 20&       &  1.03  &  1.62   &  23.55    & 201  & 216\\
                & 30&       &  1.21  &  1.88   &  24.24    & 251  & 315\\
                & 50&       &  1.53  &  2.32   &  25       & 251  & 476\\
\hline
\end{tabular}
}
\end{center}
\caption{\diff{Total runtime
		in seconds of the Trim Random, Split
		Barycenter and No Trim method on the \banosdataset dataset. The
		runtime varies depending on the number of
		trees used and the amount of memory available.}}
\label{tab:runtime-xp2}
\end{table}

\section{Related Work}
\label{sec:related-work}
Edge computing is a concept where processing is done close to the device
that produced the data, which generally means on devices with much less
memory than regular computing servers. There
are many surveys about classification for edge
computing~\cite{survey-edge-machine-learning,tinyML-survey-1, tinyML-survey-2},
but most of the work focuses on deep learning, which is not applicable in our
case because it requires a lot of data and time to train the model. They discuss
inherent problems related to learning with edge devices, in particular about
lighter architecture and distributed training. They also depict areas where
machine learning on edge devices would be impactful like computer vision, fraud
detection, or autonomous vehicles. Finally, these studies draw future work
opportunities such as data augmentation, distributed training, and explainable
AI. Aside from the deep learning approaches, the survey
in~\cite{survey-edge-machine-learning} discusses two machine learning techniques
with a small memory footprint: the Bonsai and ProtoNN methods.

Bonsai~\cite{bonzai_tree} is a tree-based algorithm designed to fit in an
edge device memory. It is a sparse tree that comes with a low-dimension
projection of the feature space to improve learning while limiting memory usage
and achieving state-of-the-art accuracy. Similarly, ProtoNN~\cite{protoNN} is a
kNN based model that performs a low-dimension projection of the features to
increase accuracy and improve its memory footprint. It also compresses the
training set into a fixed amount of clusters. ProtoNN and Bonsai claim to remain
below 2KB while retaining high accuracy. However, these models don't apply to
evolving data streams because their low-dimension projections and their
structures are pre-trained based on existing data, thus, adjusting them would
require more time and memory. Additionally, our method starts from scratch
whereas ProtoNN or Bonsai require data before being used.

When it comes to forests designed for concept drift, there are many variations
and many mechanisms. The Hoeffding Adaptive Tree~\cite{HAT} embeds a concept
drift detector and grows a ghost branch when it detects a drift in a branch. The
ghost branch replaces the old one when its performance becomes better.
Similarly, the Adaptive Random Forest~\cite{adaptive_learning_from_stream} keeps
a drift detector for each tree and starts growing a ghost tree when a drift is
detected. The work in~\cite{online_random_forest_2} presents a forest that is
constantly evolving where each decision tree has its size limit. When the limit
is reached, it restarts from the last created node. With this mechanism, trees
with a smaller limit will adapt faster to recent data points. Additionally, the
forest also embeds the ADWIN~\cite{adwin} concept drift detector and restarts the
worst base learner when a drift is detected. This mechanism called the ADWIN
bagging is used in the Adaptive Rotation Forest~\cite{adaptive_rotation_forest}
in addition to a low-dimension projection of the features with an incremental
PCA. Such combinations allow the forest to maintain the most accurate base
classifiers while keeping the projection up-to-date. Our methods differ from
ADWIN or the Adaptive Rotation Forest because we rely on passive drift
adaptation rather than using a drift detector. 

The Kappa Updated Ensemble~\cite{kappa_updated_ensemble} is an ensemble method
that notices drifts by self-monitoring the performance of its base classifiers.
In case of a drift, the model trains new classifiers. The prediction is made
using only the best classifiers from the ensemble but the method never discards
a base classifier as it can still be useful in the future. This mechanism of
keeping unused trees raises a memory problem since it may fill the memory faster
for very little benefit.

Similarly, the work in~\cite{pruning_ensemble_for_evolving_streams} proposes a
method to prune base learners based on their global and class-wise performances.
It is used to reduce memory consumption while retaining good accuracy across
all classes. The method evaluates the base learners for each class then ranks
them using a modified version Borda Count. 

The Mean error rate Weighted Online
Boosting~\cite{mean_error_weighted_online_boosting} is an online boosting method
where the weights are calculated based on the accuracy of previous data. Even
though the method is not designed for concept drift, the self-monitoring of the
accuracy makes the base learner train more on recent data making the ensemble
robust to concept drift. 

\diff{The Robust Online Self-Adjusting Ensemble
(ROSE)~\cite{rose} is a classifier ensemble
designed for imbalanced data streams with concept
drifts, where each sub-classifier is constructed
using a random feature subspace with an ADWIN
drift detector. In case a drift is detected, a
background ensemble is trained on a class-wise
sliding window, which maintains a sliding window
per class thus downsampling the majority classes.
After processing a thousand data points, the top
sub-classifiers from both the current ensemble and
the background ensemble are compared, and the best
ones are selected to form the new ensemble. This
process ensure the background ensemble is trained
on fresh and balanced data points.}

\diff{The One-Class Drift Detector~(OCDD) uses an additional
classifier to recognize a concepts. The
classifier is trained on the first complete
sliding window. From this first window, the
classifier learns the distribution of the current
concept and a drift is detected if the new data
points contains too many outliers.}

These last studies rely on ranking the base learners of the ensemble to either
adjust or disable them. However, these coarse-grain approaches are memory
intensive and are not applicable to the trimming methods because they would
require keeping statistics for each node.

\section{Conclusion}
We adapted \mondrianforests to support data streams and we proposed
 five out-of-memory strategies to deploy them under memory constraints.
Results show that the Extend Node method has the best improvement on
average. With a carefully tuned number of trees, the Extend Node also has
the highest F1 score gain compared to the Stopped strategy. Thus, we
recommend using Extend Node as the default strategy.

We also compared node trimming methods for the \mondriantrees and there are two
viable methods depending on the situation. Not trimming is the best option in
most case of stable dataset. However, when expecting a concept drift, the trim
Random with splits is preferable. The drawback with the Trim Random method is
that it deteriorates the F1 score on stable streams.

Overall, this paper showed that using our out-of-memory strategies is critical
in order to make the \mondrianforest work in a memory-constrained environment.
In particular, existing \mondrianforest
implementations~\cite{mondrian_implementation_1, mondrian_implementation_2} do
not have any out-of-memory strategy and will fail if they cannot allocate any
more nodes. Using the Extend Node strategy allows an average F1 score gain of
0.28 accross all datasets compared to not doing anything. Similarly, using Trim
Random offers an average F1 score gain of 0.3.

Our results show no significant difference between the two types of node
splitting strategy. By default, we would recommend using Split AVG because it is
less compute-intensive.

\diff{Future work will investigate trim
mechanisms to adaptively trim depending on the memory limit and concept drift.
Finally, we suggest exploring the use of drift detectors such as
ADWIN~\cite{adwin} to switch between No Trim and
Trim Random.}

\section*{Compliance with Ethical Standards}
This work was funded by a Strategic Project Grant of the Natural Sciences
and Engineering Research Council of Canada. The computing platform was
obtained with funding from the Canada Foundation
for Innovation. The authors have no conflicts of
interest.



\end{document}